\RequirePackage{amsmath}
\newcommand{\Head}[1]{\ensuremath{\mathit{H}(#1)}}
\newcommand{\body}[1]{\ensuremath{\mathit{B}(#1)}} % {\ensuremath{\mathit{body}(#1)}}

\input{./include/asp-macros/dtel}
\providecommand{\logfont}{\textrm}

\newcommand{\HT}{\ensuremath{\logfont{HT}}}

\newcommand{\LTL}{\ensuremath{\logfont{LTL}}}
\newcommand{\LTLf}{\ensuremath{\LTL_{\!f}}}

\newcommand{\THT}{\ensuremath{\logfont{THT}}}
\newcommand{\THTf}{\ensuremath{\THT_{\!f}}}

\newcommand{\TEL}{\ensuremath{\logfont{TEL}}}
\newcommand{\TELf}{\ensuremath{\TEL_{\!f}}}

\providecommand{\sysfont}{\textit}

\newcommand{\telingo}{\sysfont{telingo}}

\newcommand{\tuple}[1]{\ensuremath{\langle #1 \rangle}}
\newcommand{\Htrace}{\ensuremath{\mathbf{H}}}
\newcommand{\Ttrace}{\ensuremath{\mathbf{T}}}
\newcommand{\M}{\ensuremath{\mathbf{M}}}

\newcommand{\Bd}{\ensuremath{B}} % \def\Bd{{\cal B}}
\newcommand{\Hd}{\ensuremath{H}} % \def\Hd{{\cal H}}

\newcommand{\initial}[1]{\ensuremath{\mathit{I}(#1)}} % {\ensuremath{\mathit{Ini}(#1)}}
\newcommand{\dynamic}[1]{\ensuremath{\mathit{D}(#1)}} % {\ensuremath{\mathit{Dyn}(#1)}}
\newcommand{\final}[1]{\ensuremath{\mathit{F}(#1)}} % {\ensuremath{\mathit{Fin}(#1)}}

\newcommand{\inst}[2]{\ensuremath{#1(#2)}}

\newcommand{\eqdef}{\ensuremath{\mathbin{\raisebox{-1pt}[-3pt][0pt]{$\stackrel{\mathit{def}}{=}$}}}}

\newenvironment{proofof}[1]{\noindent {\bf Proof of #1.}}{\leavevmode\qed}
\newcommand{\A}{\ensuremath{\mathcal{A}}}

\newcommand{\tCF}[1]{\ensuremath{CF(#1)}}
\newcommand{\tCFa}[2]{\ensuremath{CF_#1(#2)}}

\newcommand{\X}{\ensuremath{\mathbf{X}}}

\newcommand{\LF}[1]{\ensuremath{\mathit{LF}(#1)}}
\newcommand{\Graph}[1]{\ensuremath{\mathit{G}(#1)}}
\newcommand{\Loops}[1]{\ensuremath{\mathit{L}(#1)}}
\newcommand{\support}[2]{\ensuremath{\mathit{S}_{#2}(#1)}}
\newcommand{\myatom}{\ensuremath{p}}

\newcommand{\T}{\ensuremath{\mathbf{T}}}

\newcommand{\intervo}[2]{[#1..#2)}
\newcommand{\rangeo}[3]{#1 \in \intervo{#2}{#3}}
\newcommand{\intervc}[2]{[#1..#2]}
\newcommand{\rangec}[3]{#1 \in \intervc{#2}{#3}}
\newcommand{\TS}{\ensuremath{\mathrm{TS}}}

\documentclass[a4paper]{include/latex-class-llncs/llncs}
\renewcommand{\next}{\text{\rm \raisebox{-.5pt}{\Large\textopenbullet}}}  % {{\ensuremath{\circ}}}
\usepackage[utf8]{inputenc}
\usepackage[inline]{enumitem}
\usepackage{amssymb}
\usepackage{multicol}
\usepackage{url}

\begin{document}

\title{Past-present temporal programs\\ over finite traces}
\author{Pedro Cabalar\inst{1}\orcidID{0000-0001-7440-0953} \and
	Mart\'in Di\'eguez\inst{2}\orcidID{0000-0003-3440-4348} \and
	Fran\c{c}ois Laferri\`ere\inst{3}\orcidID{0009-0006-8147-572X} \and
	Torsten Schaub\inst{3}\orcidID{0000-0002-7456-041X}
}
\authorrunning{Cabalar et al.}

\institute{
$^1$ University of Corunna, Spain\\$^2$ University of Angers, France\\$^3$ University of Potsdam, Germany
}

\maketitle
\begin{abstract}
Extensions of Answer Set Programming with language constructs from temporal logics,
such as temporal equilibrium logic over finite traces (\TELf), provide an expressive
computational framework for modeling dynamic applications.
In this paper, we study the so-called past-present syntactic subclass, which consists
of a set of logic programming rules whose body references to the past and head to
the present. Such restriction ensures that the past remains independent of the future,
which is the case in most dynamic domains.
We extend the definitions of completion and loop formulas to the case of past-present
formulas, which allows for capturing the temporal stable models of past-present temporal programs by
means of an \LTLf~expression.
\end{abstract}

\section{Introduction}\label{sec:introduction}

Reasoning about dynamic scenarios is a central problem in the areas of  Knowledge Representation~\cite{bale04} (KR) and Artificial Intelligence (AI).
Several formal approaches and systems have emerged to introduce non-monotonic reasoning features in scenarios where the formalisation of time is fundamental~\cite{BaralZ07,BaralZ08,emerson90a,Gonzalez2002,sandewall94a}.
In \emph{Answer Set Programming}~\cite{breitr11a} (ASP), former approaches to temporal reasoning use first-order encodings~\cite{lifschitz99b} where the time is represented by means of a variable whose value comes from a finite domain.
The main advantage of those approaches is that the computation of answer sets can be achieved via incremental solving~\cite{gekakaosscth08a}.
Their downside is that they require an explicit representation of time points.

\emph{Temporal Equilibrium Logic}~\cite{AguadoCDPSSV23} (\TEL{}) was proposed as a temporal extension of \emph{Equilibrium Logic}~\cite{pearce96a}
with connectives from \emph{Linear Time Temporal Logic}~\cite{pnueli77a} (\LTL{}).
Due to the computational complexity of its satisfiability problem (\textsc{ExpSpace}), finding tractable fragments of \TEL{} with good computational properties have also been a topic in the literature.
Within this context, \emph{splittable temporal logic programs}~\cite{agcapevi11a} have been proved to be a syntactic fragment of \TEL{} that allows for a reduction to \LTL{} via the use of Loop Formulas~\cite{feleli06a}.

When considering incremental solving, logics on finite traces such as \LTLf{}~\cite{giavar13a} have been shown to be more suitable.
Accordingly, \emph{Temporal Equilibrium Logic on Finite traces} (\TELf)~\cite{cakascsc18a} was created and became the foundations of the temporal ASP solver \telingo{}~\cite{cakamosc19a}.

We present a new syntactic fragment of \TELf{}, named \emph{past-present} temporal logic programs.
Inspired by Gabbay's seminal paper~\cite{gabbay87a}, where the declarative character of past temporal operators is emphasized, this language consists of a set of logic programming rules whose formulas in the head are disjunctions of atoms that reference the present, while in its body we allow for any arbitrary temporal formula without the use of future operators.
Such restriction ensures that the past remains independent of the future, which is the case in most dynamic domains, and makes this fragment advantageous for incremental solving.
%The use of only past operators in the body has the advantage that, when using incremental solving, the body of each rule can be seen as a query whose satisfiability can be checked on the (partial) incremental computation at each step.
%This advantage allows us to exploit the advantages of \telingo's API in order to reuse partial computations during the solving in order increase its performance.

As a contribution, we study the Lin-Zhao theorem~\cite{linjzh03a} within the context of past-present temporal logic programs.
More precisely, we show that when the program is \emph{tight}~\cite{erdlif03a}, extending Clark's completion~\cite{clark78a,fages94a} to the temporal case suffices to capture the answer sets of a finite past-present program as the \LTLf{}-models of a corresponding temporal formula.
We also show that, when the program is not tight, the use of loop formulas is necessary. To this purpose, we  extend the definition of loop formulas to the case of past-present programs and we prove the Lin-Zhao theorem in our setting.
%Finally, we also prove the generalisation of Lin-Zhao theorem in the sense of~\cite{feleli06a}, where the computation of the completion is be replaced by the consideration of unitary loops.

The paper is organised as follows: in Section~\ref{sec:background}, we review the formal background and we introduce the concept of past-present temporal programs.
In Section~\ref{sec:tcompletion}, we extend the completion property to the temporal case.
Section~\ref{sec:lf} is devoted to the introduction of our temporal extension of loop formulas.
In section~\ref{sec:unitary-cycles}, we shows that temporal completion can be captured in the general theory
of loop formulas by considering unitary cycles.
Finally, in Section~\ref{sec:conclusions}, we present the conclusions as well as some future research lines.

%After establishing the formal background
%The paper is organised as follows: in Section~\ref{sec:background} we introduce the logic of temporal here and there over finite traces and its equilibrium counterpart.
%In Section~\ref{sec:ppp} we introduce the concept of past-present temporal programs while, in Section~\ref{sec:tcompletion} we extend the completion property to the temporal case and we prove that, for the case of tight programs, computing the temporal completion suffices to characterise the temporal answer sets of a past-present programs in terms of \LTLf{} formulas.
%The non-tight case is studied in Section~\ref{sec:lf}, where we introduce our temporal extension of loop formulas and extend the results presented in Section~\ref{sec:tcompletion} to the case of non-tight programs.
%Our last contribution, presented in Section~\ref{sec:unitary-cycles}, shows that temporal completion can be captured in the general theory of loop formulas by considering unitary cycles.
%Finally, in Section~\ref{sec:conclusions} we present the conclusions of the paper an we outline some future research lines.

\section{Past-present temporal programs over finite traces}\label{sec:background}
In this section, we introduce the so-called \emph{past-present} temporal logic programs and its semantics based on \emph{Temporal Equilibrium Logic over Finite traces} (\TELf{} for short)  as in~\cite{AguadoCDPSSV23}.
The syntax of our language is inspired from the pure-past fragment of Linear Time Temporal Logic (\LTL{})~\cite{GiacomoSFR20}, since the only future operators used are \emph{always} and \emph{weak next}.

%\comment{M: rewrite}For this paper to be self-contained, we first recall the definitions of \emph{Temporal Here-and-There over finite traces} (\THTf{} for short)\ and its non-monotonic counterpart \emph{Temporal Equilibrium Logic over finite traces} (\TELf)\
%\comment{M:rewrite}All logics treated in this paper share the common syntax of \LTL{} with past operators~\cite{emerson90a}.
We start from a given set $\mathcal{A}$ of atoms which we call the \emph{alphabet}.
Then, \emph{past temporal formulas} $\varphi$ are defined by the grammar:
\[
\varphi ::= a \mid \bot\mid \neg \varphi \mid \varphi_1 \wedge \varphi_2 \mid \varphi_1 \vee \varphi_2  \mid\previous\varphi \mid \varphi_1 \since \varphi_2 \mid \varphi_1 \trigger \varphi_2
\]
where $a\in\A$ is an atom.
The intended meaning of the (modal) temporal operators is as in \LTL{}.
$\previous \varphi$  means that $\varphi$ is true at the previous time point;
$\varphi \since \psi$ can be read as $\varphi$ is true since $\psi$ was true and
$\varphi \trigger \psi$ means that $\psi$ is true since both $\varphi$ and $\psi$ became true simultaneously or $\psi$ has been true from the beginning.

Given $a \in \mathbb{N}$ and $b \in \mathbb{N}$,
we let $\intervc{a}{b}\eqdef \{i \in \mathbb{N} \mid a \leq i \leq b\}$ and
$\intervo{a}{b}\eqdef \{i \in \mathbb{N} \mid a \leq i < b\}$.
%and $\intervoc{a}{b}\eqdef \{i \in \mathbb{N} \mid a < i \leq b\}$.
%
A \emph{finite trace} \T\ of length $\lambda$ over alphabet \A\ is
a sequence $\T=(T_i)_{\rangeo{i}{0}{\lambda}}$ of sets $T_i\subseteq\A$.
%
%We say that \T\ is \emph{infinite} if $\lambda=\omega$ and \emph{finite} otherwise.
%
To represent a given trace, we write a sequence of sets of atoms concatenated with `$\cdot$'.
For instance, the finite trace $\{a\} \cdot \emptyset \cdot \{a\} \cdot \emptyset$ has length 4 and makes $a$ true at even time points and false at odd ones.
%

%At each state $T_i$ in a trace, an atom $a$  can only be true, viz.\ $a \in T_i$, or false, $a \not\in T_i$.
%%
%The logic \THT\ weakens this truth assignment, following the same intuitions as the (non-temporal) logic of \HT.
%%
%In \THT, an atom can have one of three truth-values in each state, namely,
%\emph{false}, \emph{assumed} (or true by default) or \emph{proven} (or certainly true).
%%
%Anything proved has to be assumed, but the opposite does not necessarily hold.
%%
%Following this idea,
%a state $i$ is represented as a pair of sets of atoms $\tuple{H_i,T_i}$ with $H_i \subseteq T_i \subseteq \A$ where
%$H_i$ (standing for ``here'') contains the proven atoms, whereas $T_i$ (standing for ``there'') contains the assumed atoms.
%%
%On the other hand, false atoms are just the ones not assumed, captured by $\A \setminus T_i$.
%%
%Accordingly,
A \emph{Here-and-There trace} (for short \emph{\HT-trace}) of length $\lambda$ over alphabet \A\ is
a sequence of pairs
\(
(\tuple{H_i,T_i})_{\rangeo{i}{0}{\lambda}}
\)
with $H_i\subseteq T_i$ for any $\rangeo{i}{0}{\lambda}$.
For convenience, we usually represent the \HT-trace as the pair $\tuple{\Htrace,\Ttrace}$ of traces $\Htrace = (H_i)_{\rangeo{i}{0}{\lambda}}$ and $\T = (T_i)_{\rangeo{i}{0}{\lambda}}$.
Given $\M=\tuple{\Htrace,\Ttrace}$, we also denote its length as $|\M| \eqdef |\Htrace|=|\T|=\lambda$.
Note that the two traces \Htrace, \T\ must satisfy a kind of order relation, since $H_i \subseteq T_i$ for each time point $i$.
Formally, we define the ordering $\Htrace \leq \T$ between two traces of the same length $\lambda$ as $H_i\subseteq T_i$ for each $\rangeo{i}{0}{\lambda}$.
Furthermore, we define $\Htrace<\T$ as both $\Htrace\leq\T$ and $\Htrace\neq\T$.
Thus, an \HT-trace can also be defined as any pair $\tuple{\Htrace,\Ttrace}$ of traces such that $\Htrace \leq \T$.
The particular type of \HT-traces satisfying $\Htrace=\T$ are called \emph{total}.

%We proceed by generalizing the extension of \HT\ with temporal operators,
%called \THT~\cite{agcadipevi13a},
%to \HT-traces of fixed length in order to integrate finite as well as infinite traces.
%%
%Given any \HT-trace $\M=\tuple{\Htrace,\Ttrace}$,
%we define the \THT\ satisfaction of formulas as follows.
%
% --------------------------------------------------------------------------------
%\begin{definition}[\THT-satisfaction;~\cite{agcadipevi13a,cakascsc18a}]\label{def:dht:satisfaction}%
  %\footnotetext{The while operator \while\ was introduced by~\cite{agcafapevi20a}.}
  An \HT-trace $\M=\tuple{\Htrace,\Ttrace}$ of length $\lambda$ over alphabet \A\
  \emph{satisfies} a past temporal formula $\varphi$ at time point $\rangeo{k}{0}{\lambda}$,
  written \mbox{$\M,k \models \varphi$}, if the following conditions hold:
  \begin{enumerate}
  \item $\M,k \models \top$ and  $\M,k \not\models \bot$
  \item $\M,k \models \myatom$ if $\myatom \in H_k$ for any atom $\myatom \in \A$
  \item $\M, k \models \varphi \wedge \psi$
    iff
    $\M, k \models \varphi$
    and
    $\M, k \models \psi$
  \item $\M, k \models \varphi \vee \psi$
    iff
    $\M, k \models \varphi$
    or
    $\M, k \models \psi$
  \item $\M, k \models \neg \varphi$
    iff
    $\langle \mathbf{T}, \mathbf{T} \rangle, k \not \models \varphi$
  \item $\M, k \models \previous \varphi$
    iff
    $k>0$ and $\M, k{-}1 \models \varphi$
  \item $\M, k \models \varphi \, \since \, \psi$
    iff
    for some $\rangec{j}{0}{k}$, we have
    $\M, j \models \psi$
    and
    $\M, i \models \varphi$ for all $\rangeoc{i}{j}{k}$
  \item $\M, k \models \varphi \trigger \psi$
    iff
    for all $\rangec{j}{0}{k}$, we have
    $\M, j \models \psi$
    or
    $\M, i \models \varphi$ for some $\rangeoc{i}{j}{k}$
%  \item $\M, k \models \next \varphi$
%    iff
%    $k+1<\lambda$ and $\M, k{+}1 \models \varphi$
%  \item $\M, k \models \varphi \until \psi$
%    iff
%    for some $\rangeo{j}{k}{\lambda}$, we have
%    $\M, j \models \psi$
%    and
%    $\M, i \models \varphi$ for all $\rangeo{i}{k}{j}$
%  \item $\M, k \models \varphi \release \psi$
%    iff
%    for all $\rangeo{j}{k}{\lambda}$, we have
%    $\M, j \models \psi$
%    or
%    $\M, i \models \varphi$ for some $\rangeo{i}{k}{j}$
    %
%  \item $\M, k \models \varphi \while \psi$
%    iff
%    for all $\rangeo{j}{k}{\lambda}$, we have
%    $\tuple{\Htrace',\Ttrace},j \models \varphi$ or $\tuple{\Htrace',\Ttrace},i \not\models \psi$ for some $\rangeo{i}{k}{j}$ and for all $\Htrace' \in \{\Htrace,\T\}$
  \end{enumerate}
% \qed
%\end{definition}

A formula $\varphi$ is a \emph{tautology} (or is \emph{valid}), written $\models \varphi$,
iff $\M,k \models \varphi$ for any \HT-trace \M\ and any $\rangeo{k}{0}{\lambda}$.
We call the logic induced by the set of all tautologies \emph{Temporal logic of Here-and-There over finite traces} (\THTf\ for short).

%\begin{proposition}
    The following equivalences hold in  \THTf{}:
\begin{enumerate*}
%	\item \(\neg \varphi \eqdef  \varphi \to \bot\),
	\item \(\top \equiv \neg \bot\),
	\item $\initially \equiv  \neg \previous \top$,
%	\item \(\varphi \leftrightarrow \psi \eqdef (\varphi \to \psi) \wedge (\psi \to \varphi)\),
	\item $\alwaysP \varphi  \equiv \bot \trigger \varphi$,
	\item $\eventuallyP \varphi   \equiv  \top \since \varphi$,
	\item $\wprevious \varphi   \equiv  \previous \varphi \vee \initially$.
\end{enumerate*}
\begin{definition}[Past-present Program]
	Given alphabet $\mathcal{A}$, the set of \emph{regular literals} is defined  as
	$\{a, \neg a,  \mid a \in \mathcal{A}\}$.

	A \emph{past-present rule} is either:
	\par
	\begin{tabular}{l@{\quad}r@{\;}c@{\;}l}
		\hspace{\itemindent} \ {\labelitemi} \ an \emph{initial rule} of form &                           $\Hd$ & $\leftarrow$ & $\Bd$     \\
		\hspace{\itemindent} \ {\labelitemi} \ a  \emph{dynamic rule} of form &  $\wnext\,\alwaysF (       \Hd$ & $\leftarrow$ & $\Bd)$    \\
		\hspace{\itemindent} \ {\labelitemi} \ a  \emph{final   rule} of form &  $\alwaysF(\,\finally \to (\bot$ & $\leftarrow$ & $\Bd )\,)$
	\end{tabular}
	\par
	where $\Bd$ is an pure past formula for dynamic rules and $\Bd=b_1 \wedge \dots \wedge b_n$
	with $n\geq 0$ for initial and final rules, the $b_i$ are regular literals,
	$\Hd=a_1 \vee \dots \vee a_m$ with $m \geq 0$ and $a_j\in \mathcal{A}$.
	A \emph{past-present program} is a set of past-present rules.\qed
\end{definition}

We let \initial{P}, \dynamic{P}, and \final{P} stand for the set of all initial, dynamic, and final rules in a
past-present program $P$, respectively.
Additionally we refer to \Hd\ as the \emph{head} of a rule $r$ and to \Bd\ as the \emph{body} of $r$.
We let $\body{r}=B$ and $\Head{r}=H$ for all types of rules above.
For example, let consider the following past-present program $P_1$:
\begin{align}
	&load \leftarrow  \label{r1}\\
	\wnext\alwaysF(&shoot \vee load \vee unload \leftarrow \ ) \label{r2}\\
	\wnext\alwaysF(&dead  \leftarrow shoot \wedge \neg unload \since load)\label{r3}\\
	\wnext\alwaysF(&shoot  \leftarrow dead)\label{r4}\\
	\alwaysF(&\finally \to (\bot \leftarrow\neg dead))\label{r5}
\end{align}
%The set of bodies in $P$ is $\body{P} = \{\body{r} \mid r\in P\}$.
%The set of heads in $P$ is $\Head{P} = \{\Head{r} \mid r\in P\}$.
%Rule~(\ref{r1}) is the only initial rule of $P$, rules~(\ref{r2}) to~(\ref{r4}) are dynamic rules, and rule~(\ref{r5}) is a final rule.
We get $\initial{P_1}=\{(\ref{r1})\}$, $\dynamic{P_1}=\{(\ref{r2}), (\ref{r3}), (\ref{r4})\}$, and $\final{P_1}=\{(\ref{r3})\}$.
Rule~(\ref{r1}) states that the gun is initially loaded.
Rule~(\ref{r2}) gives the choice to shoot, load, or unload the gun.
Rule~(\ref{r3}) states that if the gun is shot while it has been loaded, and not unloaded since, the target is dead.
Rule~(\ref{r4}) states that if the target is dead, we shoot it again.
Rule~(\ref{r5}) ensures that the target is dead at the end of the trace.

The satisfaction relation of a past-present rule on an \HT-trace $\M$ of length $\lambda$ and
at time point $\rangeo{k}{0}{\lambda}$ is defined below:

\begin{itemize}
\item $\M, k \models \Hd \leftarrow \Bd$ iff $\M', k \not \models \Bd$ or $\M', k \models \Hd$, for all $\M' \in \lbrace \M,\tuple{\T,\T} \rbrace$
\item $\M, k \models \wnext\,\alwaysF ( \Hd$  $\leftarrow$ $\Bd)$  iff $\M', i \not \models \Bd$ or $\M', i \models \Hd$ for all $\M' \in \lbrace \M,\tuple{\T,\T} \rbrace$ and all $\rangeo{i}{k+1}{\lambda}$
\item $\M, k \models \alwaysF(\,\finally \to (\bot \leftarrow \Bd )\,)$ iff $\tuple{\T,\T}, \lambda -1 \not \models \Bd$
\end{itemize}

An \HT-trace $\M$ is a \emph{model} of a past-present program $P$ if $\M,0 \models r$ for all rule $r \in P$.
Let $P$ be past-present program.
A total \HT-trace $\tuple{\Ttrace,\Ttrace}$ is a \emph{temporal equilibrium model} of $P$ iff
$\tuple{\Ttrace,\Ttrace}$ is a \emph{model} of $P$,
and there is no other $\Htrace < \T$ such that $\tuple{\Htrace,\Ttrace}$ is a \emph{model} of $P$.
The trace \T\ is called a \emph{temporal stable model} (\TS-model) of $P$.
% \qed
% \end{definition}

For length $\lambda = 2$, $P_1$ has a unique \TS-model $\{load\}\cdot\{shoot, dead\}$.

\section{Temporal completion}\label{sec:tcompletion}
In this section, we extend the completion property to the temporal case of past-present programs.

%\begin{definition}[Positive occurence]
An occurrence of an atom in a formula is \emph{positive} if it is in the antecedent of an even number of implications,
negative otherwise.
%\end{definition}
%\begin{definition}[Present occurence]
An occurrence of an atom in a formula is \emph{present} if it is not in the scope of $\previous$ (previous).
%\end{definition}
%\begin{definition}[Dependency graph]
Given a past-present program $P$ over $\mathcal{A}$,
we define its \emph{(positive) dependency graph}
\Graph{P} as $(\mathcal{A}, E)$ such that
$(a,b) \in E$ if there is a rule $r\in P$ such that $a\in\Head{r}\cap\mathcal{A}$ and $b$
has positive and present occurence in $\body{r}$  that is not in the scope of negation.
%\end{definition}
%\begin{definition}[Loop]
A nonempty set $L\subseteq \mathcal{A}$ of atoms is called \emph{loop} of $P$ if, for every pair $a,b$ of atoms in $L$,
there exists a path of length $> 0$ from $a$ to $b$ in \Graph{P} such that all vertices in the path belong to $L$.
%\end{definition}
We let \Loops{P} denote the set of loops of $P$.

Due to the structure of past-present programs, dependencies from the future to the past cannot happen,
and therefore there can only be loops within a same time point.
To reflect this, the definitions above only consider atoms with present occurences.
For example, rule $a\leftarrow b\wedge \previous c$ generates the edge $(a,b)$ but not $(a,c)$.

%\begin{definition}[Tight program]
%    A past-present program $P$ is said to be \emph{tight} if $\initial{P}$ and
%    $\dynamic{P}$ do not contain any loop.
%\end{definition}
For $P_1$, we get for the initial rules $\Graph{\initial{P_1}}=(\{load, unload, shoot, dead\},\emptyset)$ whose loops are $\Loops{\initial{P_1}}=\emptyset$.
For the dynamic rules, we get
$\Graph{\dynamic{P_1}}=(\{load,\linebreak[1] unload,\linebreak[1] shoot,\linebreak[1] dead\},\{(dead,shoot),\linebreak[1] (dead, load),\linebreak[1] (shoot, dead)\})$ and
$\Loops{\dynamic{P_1}}=\{\{shoot, dead\}\}$.
% \begin{align*}
%     \Graph{\dynamic{P_1}}&=(\{load, unload, shoot, dead\},\{(dead,shoot), (dead, load), (shoot, dead)\})\\
%     \Loops{\dynamic{P_1}}&=\{\{shoot, dead\}\}
% \end{align*}

In the following, $\varphi\rightarrow\psi\eqdef\psi\leftarrow\varphi$ and
$\varphi\leftrightarrow\psi\eqdef\varphi\rightarrow\psi\wedge\varphi\leftarrow\psi$.
\begin{definition}[Temporal completion]
    We define the temporal \emph{completion} formula of an atom $a$ in a past-present program $P$ over \A, denoted \tCFa{P}{a} as:

        \begin{equation*}
            \alwaysF\big( a \leftrightarrow
                \ \bigvee_{r\in\initial{P}, a\in\Head{r}} (\initially\wedge S(r,a))
            \vee \bigvee_{r\in\dynamic{P}, a\in\Head{r}} (\neg\initially\wedge S(r,a))\big)
        \end{equation*}

\noindent where  \(
  S(r,a) = \body{r}\wedge \bigwedge_{p\in \Head{r}\setminus \{a\}} \neg p
  \).

The temporal \emph{completion} formula of $P$, denoted \tCF{P}, is
    \begin{equation*}
        \{\tCFa{P}{a}	\mid a\in\A\}
        \cup \{r\mid r\in \initial{P}\cup\dynamic{P}, \Head{r}=\bot\}
        \cup \final{P}.
    \end{equation*}

\end{definition}

    A past-present program $P$ is said to be \emph{tight} if $\initial{P}$ and
    $\dynamic{P}$ do not contain any loop.
\begin{theorem}\label{thm:tcompletion}
    Let $P$ be a tight past-present program and \T\ a trace of length $\lambda$.
    Then, \T\ is a \TS-model of $P$ iff \T\ is a \LTLf-model of \tCF{P}.\qed
\end{theorem}

The completion of $P_1$ is
\begin{equation*}
\tCF{P_1} = \left\{
\begin{array}{c}
\alwaysF(load \leftrightarrow \initially \vee
(\neg\initially\wedge \neg shoot \wedge \neg unload)),\\
\alwaysF(shoot \leftrightarrow (\neg\initially\wedge \neg load \wedge \neg unload))
\vee (\neg\initially\wedge dead)), \\
\alwaysF(unload \leftrightarrow (\neg\initially\wedge \neg shoot \wedge \neg load)), \\
\alwaysF(dead \leftrightarrow (\neg\initially\wedge shoot \wedge \neg unload \since load)), \\
\alwaysF(\finally \to (\bot \leftarrow\neg dead))
\end{array}
\right\}.
\end{equation*}

For $\lambda=2$, $\tCF{P_1}$ has a unique \LTLf-model $\{load\}\cdot\{shoot, dead\}$,
which is identical to the \TS-model of $P_1$.
Notice that for this example, the \TS-models of the program match the \LTLf-models
of its completion despite the program not being tight. This is generally not the case.
Let $P_2$ be the program made of the rules $(\ref{r1}), (\ref{r3}), (\ref{r4})$ and $(\ref{r5})$.
The completion of $P_2$ is
\begin{equation*}
\tCF{P_2} = \left\{
\begin{array}{ll}
& \alwaysF(load \leftrightarrow \initially),\;
\alwaysF(shoot \leftrightarrow (\neg\initially\wedge dead)), \;
\alwaysF(unload \leftrightarrow \bot),\\
&\alwaysF(dead \leftrightarrow (\neg\initially\wedge shoot \wedge \neg unload \since load)),\; \alwaysF(\finally \to (\bot \leftarrow\neg dead))
\end{array}
	\right\}.
\end{equation*}

$P_2$ does not have any \TS-model, but $\{load\}\cdot\{shoot, dead\}$ is a \LTLf-model of $\tCF{P_2}$.
Under ASP semantics, it is impossible to derive any element of the loop $\{shoot, dead\}$,
as deriving $dead$ requires $shoot$ to be true, and deriving $shoot$ requires $dead$ to be true.
The completion does not restrict this kind of circular derivation and therefore is insufficient to fully capture ASP semantics.

\section{Temporal loop formulas}\label{sec:lf}
To restrict circular derivations, Lin and Zhao introduced the concept of loop formulas in~\cite{linjzh03a}.
In this section, we extend their work to past-present programs.
\begin{definition}
    Let $\varphi$ be a implication-free past-present formula and $L$ a loop.
    We define the supporting transformation of $\varphi$ with respect to $L$ as
\begin{eqnarray*}
	\support{L}{\bot} & \eqdef & \bot \\
	% \support{L}{\myatom} & \eqdef &
	% \begin{cases}
	% 	\bot & \text{if} \ \myatom \in L  \\
	% 	\myatom & \text{otherwise}
	% \end{cases}\hspace{15pt} \text{for any atom } \myatom \in \A\\
    \support{L}{\myatom}  &\eqdef &\bot \text{ if } \myatom \in L\text{ ; }
    \myatom \text{ otherwise, for any } \myatom \in \A\\
	\support{L}{\neg\varphi} & \eqdef & \neg\varphi \\
	\support{L}{\varphi\wedge\psi} & \eqdef & \support{L}{\varphi}\wedge\support{L}{\psi} \\
	\support{L}{\varphi\vee\psi} & \eqdef & \support{L}{\varphi}\vee\support{L}{\psi} \\
	\support{L}{\previous\varphi} & \eqdef & \previous\varphi\\
    \support{L}{\varphi\trigger\psi} & \eqdef & \support{L}{\psi}\wedge
                                            (\support{L}{\varphi}\vee\previous(\varphi\trigger\psi))\\
    \support{L}{\varphi\since\psi} & \eqdef & \support{L}{\psi}\vee
                                            (\support{L}{\varphi}\wedge\previous(\varphi\since\psi))
\end{eqnarray*}\qed
%\begin{itemize}[itemsep=0pt]
%	\item $\support{L}{\myatom}  \eqdef \bot$ if $\myatom \in L$; $\myatom$ otherwise, for any $\myatom \in \A$; \hspace{4pt} $\support{L}{\bot}  \eqdef  \bot$; \hspace{4pt}  $\support{L}{\neg\varphi}  \eqdef  \neg\varphi$;
%	\item  $\support{L}{\varphi\wedge\psi}  \eqdef  \support{L}{\varphi}\wedge\support{L}{\psi}$; \hspace{5pt} $\support{L}{\varphi\vee\psi}  \eqdef  \support{L}{\varphi}\vee\support{L}{\psi}$; \hspace{5pt} $\support{L}{\previous\varphi}  \eqdef  \previous\varphi$ ;
%	\item  $\support{L}{\varphi\trigger\psi}  \eqdef  \support{L}{\psi}\wedge (\support{L}{\varphi}\vee\previous(\varphi\trigger\psi))$; \hspace{5pt} $\support{L}{\varphi\since\psi}  \eqdef  \support{L}{\psi}\vee
%	(\support{L}{\varphi}\wedge\previous(\varphi\since\psi))$.
%\end{itemize}\qed

%\begin{tabular}{lll}
%
%
%\multicolumn{2}{l}{

%\end{tabular}
%\begin{cases}
%	\bot & \text{if} \ \myatom \in L  \\
%	\myatom & \text{otherwise}
%\end{cases}\hspace{5pt} \text{with } \myatom \in \A$  \\[20pt]
%
% &  \\
%
\end{definition}

\begin{definition}[External support]
Given a past-present program $P$,
the external support formula of a set of atoms $L \subseteq \mathcal{A}$ wrt $P$, is defined as
\begin{equation*}
    \mathit{ES}_P(L) = \bigvee_{r\in P,\Head{r} \cap L \neq \emptyset } \big ( \support{L}{\body{r}} \wedge \bigwedge_{a\in \Head{r}\setminus L} \neg a \big)
\end{equation*}\qed
\end{definition}

For instance, for $L=\{shoot, dead\}$, $\mathit{ES}_{P_2}(L)$ and $\mathit{ES}_{P_1}(L)$ are
\begin{align*}
    \mathit{ES}_{P_2}(L) &= \support{L}{dead} \vee \support{L}{shoot \wedge \neg unload\hspace{1pt}\since\hspace{1pt} load}\\
                         &= \support{L}{dead} \vee (\support{L}{shoot}\wedge\support{L}{ \neg unload\hspace{1pt}\since\hspace{1pt} load})\\
                         &= \support{L}{dead} \vee (\support{L}{shoot} \wedge \support{L}{\neg unload} \vee \previous(\neg unload \since load))\\
                         &= \bot \vee (\bot \wedge \neg unload \vee \previous(\neg unload \since load)) = \bot.\\
    \mathit{ES}_{P_1}(L) &= \support{L}{dead} \vee \support{L}{shoot \wedge \neg unload\hspace{1pt}\since\hspace{1pt} load} \vee (\neg load \wedge \neg unload)\\
						 &= \neg load \wedge \neg unload.
\end{align*}
Rule~(\ref{r2}) provides an external support for $L$. The body $dead$ of rule~(\ref{r4}) is also a support for $L$, but not external as $dead$ belongs to $L$.
The supporting transformation only keeps external supports by removing from the body any positive and present occurence of element of $L$.

\begin{definition}[Loop formulas]
We define the set of loop formulas of a past-present program $P$ over $\mathcal{A}$, denoted \LF{P}, as:
  \begin{align*}
    \bigvee_{a\in L} a \rightarrow \mathit{ES}_{\initial{P}}(L)
            & \text{ for any loop } L \text{ in } \initial{P}\\
    \wnext \alwaysF \Big(\bigvee_{a\in L} a \rightarrow \mathit{ES}_{\dynamic{P}}(L) \Big)
            & \text{ for any loop } L \text{ in } \dynamic{P}
  \end{align*}
\end{definition}
\begin{theorem}\label{thm:loops}
  Let $P$ be a past-present program and $\T$ a trace of length $\lambda$.
  Then,
  \T\ is a \TS-model of $P$
  iff
  \T\ is a \LTLf-model of $\tCF{P} \cup \LF{P}$.\qed
\end{theorem}

For our examples, we have that
$\LF{P_1}= \wnext\alwaysF(shoot\vee dead \rightarrow \neg load \wedge\neg unload)$ and
$\LF{P_2}= \wnext\alwaysF(shoot\vee dead \rightarrow \bot)$.
It can be also checked that $\{load\}\cdot\{shoot, dead\}$ satisfies $\LF{P_1}$, but not $\LF{P_2}$.
So, we have that $\tCF{P_1} \cup \LF{P_1}$ has a unique \LTLf-model $\{load\}\cdot\{shoot, dead\}$, while
$\tCF{P_2} \cup \LF{P_2}$ has no \LTLf-model, matching the \TS-models of the respective programs.

% Ferraris et al.~\cite{feleli06a} proposed an approach where the computation of the completion
% can be avoided by considering unitary cycles.
% We extended such results for past-present programs in the extended version~\cite{cabalar2023pastpresent}.

%%% Local Variables:
%%% mode: latex
%%% TeX-master: "paper"
%%% End:

\section{Temporal loop formulas with unitary cycles}\label{sec:unitary-cycles}

Ferraris et al.~\cite{feleli06a} proposed an approach where the computation of the completion
can be avoided by considering unitary cycles.
In this section, we extend such results for past-present programs.
We first redefine loops so that unitary cycles are included.
\begin{definition}[Unitary cycle]
  A nonempty set $L\subseteq \mathcal{A}$ of atoms is called $\emph{loop}$ of $P$ if, for every pair $a,b$ of atoms in $L$,
  there exists a path (possibly of length 0) from $a$ to $b$ in $G(P)$ such that all vertices in the path belong to $L$.
\end{definition}

With this definition, it is clear that any set consisting of a single atom is a loop.
For example, $\Loops{\dynamic{P_1}}=\{\{load\},\{unload\},\{shoot\},\{dead\},\{shoot, dead\}\}$.

\begin{theorem}\label{thm:loops0}
  Let  $P$ be a past-present program and $\T$ a trace of length $\lambda$.
  Then,
  \T\ is a \TS-model of $P$
  iff
  \T\ is a \LTLf-model of $P \cup \LF{P}$.
\end{theorem}

With unitary cycle, $\LF{P_1}$ becomes
\begin{align*}
    load &\rightarrow \top\\
    unload &\rightarrow \bot\\
    shoot &\rightarrow \bot\\
    dead &\rightarrow \bot\\
    \wnext\alwaysF( load &\rightarrow \neg shoot \wedge\neg unload)\\
    \wnext\alwaysF( unload &\rightarrow \neg shoot \wedge\neg load)\\
    \wnext\alwaysF( shoot &\rightarrow (\neg shoot \wedge\neg load)\vee dead)\\
    \wnext\alwaysF( dead &\rightarrow shoot \wedge \neg unload \since load)\\
    \wnext\alwaysF(shoot\vee dead &\rightarrow \neg load \wedge\neg unload)
\end{align*}

$P_1\cup\LF{P_1}$ has the same \LTLf-model $\tCF{P_1} \cup \LF{P_1}$, $\{load\}\cdot\{shoot, dead\}$, which is the \TS-model of $P_1$.

\section{Conclusion}\label{sec:conclusions}

We have focused on temporal logic programming within the context of Temporal Equilibrium Logic over finite traces.
More precisely, we have studied a fragment close to logic programming rules in the spirit of~\cite{gabbay87a}: a past-present temporal logic program
consists of a set of rules whose body refers to the past and present while their head refers to the present.
This fragment is very interesting for implementation purposes since it can be solved by means of incremental solving techniques as implemented in \telingo{}.
%Moreover, restricting the body of the rules to only present and past formulas makes it so that the body of the rules can be seen as queries on the set of conclusions generated during the solving phase.

Contrary to the propositional case~\cite{feleli06a}, where answer sets of an arbitrary propositional formula can be captured by means of the classical models of another formula $\psi$, in the temporal case, this is impossible to do the same mapping among the temporal equilibrium models of a formula $\varphi$ and the \LTL{} models of another formula $\psi$~\cite{bozpea16a}.

In this paper, we show that past-present temporal logic programs can be effectively reduced to \LTL{} formulas by means of completion and loop formulas.
More precisely, we extend the definition of completion and temporal loop formulas in the spirit of Lin and Zhao~\cite{linjzh03a} to the temporal case, and we show that for tight past-present programs, the use of completion is sufficient to achieve a reduction to an \LTLf{} formula.
Moreover, when the program is not tight, we also show that the computation of the temporal completion and a finite number of loop formulas suffices to reduce \TELf{} to \LTLf{}.
%Finally, we consider Ferraris et al. approach~\cite{feleli06a} where the computation of the completion can be automatically replaced by the consideration of unitary loops.
%In this contribution, we extend such result to the case of past-present logic programs.

%As future work, we plan to study in detail the relation between temporal completion and loop formulas and \emph{unfounded sets}~\cite{feleli06a}, since the latter play a central role in the solving algorithm of \clingo{}~\cite{gekakasc14a}.
%Lastly, we will study the case of past-present temporal programs with variables~\cite{AguadoCPVD17} in order to get a second-order characterisation of loop formulas, in the spirit of~\cite{leemen08a}.

\subsubsection*{Acknowledgments}
This work was supported by MICINN, Spain, grant PID2020-116201GB-I00, Xunta de Galicia, Spain (GPC ED431B
2019/03), R{\'e}gion Pays de la Loire, France, (project etoiles montantes CTASP) and DFG grant SCHA 550/15, Germany.

%%% Local Variables:
%%% mode: latex
%%% TeX-master: "paper"
%%% End:

\bibliographystyle{include/latex-class-llncs/splncs04}
\bibliography{krr,procs,local}

\begin{thebibliography}{10}
\providecommand{\url}[1]{\texttt{#1}}
\providecommand{\urlprefix}{URL }
\providecommand{\doi}[1]{https://doi.org/#1}

\bibitem{agcapevi11a}
Aguado, F., Cabalar, P., P{\'{e}}rez, G., Vidal, C.: Loop formulas for
  splitable temporal logic programs. In: Delgrande, J., Faber, W. (eds.)
  Proceedings of the Eleventh International Conference on Logic Programming and
  Nonmonotonic Reasoning (LPNMR'11). Lecture Notes in Artificial Intelligence,
  vol.~6645, pp. 80--92. Springer-Verlag (2011)

\bibitem{AguadoCDPSSV23}
Aguado, F., Cabalar, P., Di{\'{e}}guez, M., P{\'{e}}rez, G., Schaub, T.,
  Schuhmann, A., Vidal, C.: Linear-time temporal answer set programming. Theory
  Pract. Log. Program.  \textbf{23}(1),  2--56 (2023)

\bibitem{BaralZ07}
Baral, C., Zhao, J.: Non-monotonic temporal logics for goal specification. In:
  Veloso, M.M. (ed.) {IJCAI} 2007, Proceedings of the 20th International Joint
  Conference on Artificial Intelligence, Hyderabad, India, January 6-12, 2007.
  pp. 236--242 (2007)

\bibitem{BaralZ08}
Baral, C., Zhao, J.: Non-monotonic temporal logics that facilitate elaboration
  tolerant revision of goals. In: Fox, D., Gomes, C.P. (eds.) Proceedings of
  the Twenty-Third {AAAI} Conference on Artificial Intelligence, {AAAI} 2008,
  Chicago, Illinois, USA, July 13-17, 2008. pp. 406--411. {AAAI} Press (2008)

\bibitem{bozpea16a}
Bozzelli, L., Pearce, D.: On the expressiveness of temporal equilibrium logic.
  In: Michael, L., Kakas, A. (eds.) Proceedings of the Fifteenth European
  Conference on Logics in Artificial Intelligence (JELIA'16). Lecture Notes in
  Artificial Intelligence, vol. 10021, pp. 159--173. Springer-Verlag (2016)

\bibitem{bale04}
Brachman, R.J., Levesque, H.J.: Knowledge Representation and Reasoning.
  Elsevier (2004),
  \url{http://www.elsevier.com/wps/find/bookdescription.cws\_home/702602/description}

\bibitem{breitr11a}
Brewka, G., Eiter, T., Truszczy{\'n}ski, M.: Answer set programming at a
  glance. Communications of the {ACM}  \textbf{54}(12),  92--103 (2011)

\bibitem{cakamosc19a}
Cabalar, P., Kaminski, R., Morkisch, P., Schaub, T.: telingo = {ASP} + time.
  In: Balduccini, M., Lierler, Y., Woltran, S. (eds.) Proceedings of the
  Fifteenth International Conference on Logic Programming and Nonmonotonic
  Reasoning (LPNMR'19). Lecture Notes in Artificial Intelligence, vol. 11481,
  pp. 256--269. Springer-Verlag (2019)

\bibitem{cakascsc18a}
Cabalar, P., Kaminski, R., Schaub, T., Schuhmann, A.: Temporal answer set
  programming on finite traces. Theory and Practice of Logic Programming
  \textbf{18}(3-4),  406--420 (2018)

\bibitem{clark78a}
Clark, K.: Negation as failure. In: Gallaire, H., Minker, J. (eds.) Logic and
  Data Bases, pp. 293--322. Plenum Press (1978)

\bibitem{giavar13a}
{De Giacomo}, G., Vardi, M.: Linear temporal logic and linear dynamic logic on
  finite traces. In: Rossi, F. (ed.) Proceedings of the Twenty-third
  International Joint Conference on Artificial Intelligence (IJCAI'13). pp.
  854--860. IJCAI/AAAI Press (2013)

\bibitem{emerson90a}
Emerson, E.: Temporal and modal logic. In: van Leeuwen, J. (ed.) Handbook of
  Theoretical Computer Science, pp. 995--1072. MIT Press (1990)

\bibitem{erdlif03a}
Erdem, E., Lifschitz, V.: Tight logic programs. Theory and Practice of Logic
  Programming  \textbf{3}(4-5),  499--518 (2003)

\bibitem{fages94a}
Fages, F.: Consistency of {C}lark's completion and the existence of stable
  models. Journal of Methods of Logic in Computer Science  \textbf{1},  51--60
  (1994)

\bibitem{feleli06a}
Ferraris, P., Lee, J., Lifschitz, V.: A generalization of the {L}in-{Z}hao
  theorem. Annals of Mathematics and Artificial Intelligence  \textbf{47}(1-2),
   79--101 (2006)

\bibitem{gabbay87a}
Gabbay, D.: The declarative past and imperative future: Executable temporal
  logic for interactive systems. In: Banieqbal, B., Barringer, H., Pnueli, A.
  (eds.) Proceedings of the Conference on Temporal Logic in Specification.
  Lecture Notes in Computer Science, vol.~398, pp. 409--448. Springer-Verlag
  (1987)

\bibitem{gekakaosscth08a}
Gebser, M., Kaminski, R., Kaufmann, B., Ostrowski, M., Schaub, T., Thiele, S.:
  Engineering an incremental {ASP} solver. In: {Garcia de la Banda}, M.,
  Pontelli, E. (eds.) Proceedings of the Twenty-fourth International Conference
  on Logic Programming (ICLP'08). Lecture Notes in Computer Science, vol.~5366,
  pp. 190--205. Springer-Verlag (2008)

\bibitem{GiacomoSFR20}
Giacomo, G.D., Stasio, A.D., Fuggitti, F., Rubin, S.: Pure-past linear temporal
  and dynamic logic on finite traces. In: Bessiere, C. (ed.) Proceedings of the
  Twenty-ninth International Joint Conference on Artificial Intelligence,
  (IJCAI'20). pp. 4959--4965. ijcai.org (2020)

\bibitem{Gonzalez2002}
Gonz{\'a}lez, G., Baral, C., Cooper, P.A.: Modeling Multimedia Displays Using
  Action Based Temporal Logic, pp. 141--155. Springer US, Boston, MA (2002)

\bibitem{lifschitz99b}
Lifschitz, V.: Answer set planning. In: {de Schreye}, D. (ed.) Proceedings of
  the International Conference on Logic Programming (ICLP'99). pp. 23--37. MIT
  Press (1999)

\bibitem{linjzh03a}
Lin, F., Zhao, J.: On tight logic programs and yet another translation from
  normal logic programs to propositional logic. In: Gottlob, G., Walsh, T.
  (eds.) Proceedings of the Eighteenth International Joint Conference on
  Artificial Intelligence (IJCAI'03). pp. 853--858. Morgan Kaufmann Publishers
  (2003)

\bibitem{pearce96a}
Pearce, D.: A new logical characterisation of stable models and answer sets.
  In: Dix, J., Pereira, L., Przymusinski, T. (eds.) Proceedings of the Sixth
  International Workshop on Non-Monotonic Extensions of Logic Programming
  (NMELP'96). Lecture Notes in Computer Science, vol.~1216, pp. 57--70.
  Springer-Verlag (1997)

\bibitem{pnueli77a}
Pnueli, A.: The temporal logic of programs. In: Proceedings of the Eight-teenth
  Symposium on Foundations of Computer Science (FOCS'77). pp. 46--57. {IEEE}
  Computer Society Press (1977)

\bibitem{sandewall94a}
Sandewall, E.: Features and fluents: the representation of knowledge about
  dynamical systems, vol.~1. Oxford University Press, New York, NY, USA (1994)

\end{thebibliography}

\appendix
\section{Proofs}

\begin{definition}\label{def:forgettable:future}
Let $\tuple{\Htrace,\Ttrace}$ and $\tuple{\Htrace',\Ttrace}$ be two \HT-traces of length $\lambda$ and let $\rangeco{i}{0}{\lambda}$. We say denote by
$\tuple{\Htrace,\Ttrace} \sim_{i}\tuple{\Htrace',\Ttrace}$ the fact for all $\rangeo{j}{0}{i}$, $H_i = H'_i$.
\end{definition}

\begin{proposition}\label{prop:pure-past}
For all \HT-traces $\tuple{\Htrace,\Ttrace}$ and $\tuple{\Htrace',\Ttrace}$  and for all $\rangeco{i}{0}{\lambda}$, if
$\tuple{\Htrace,\Ttrace} \sim_{i}\tuple{\Htrace',\Ttrace}$ then for all $\rangeo{j}{0}{i}$ and for all past formulas $\varphi$,
$\tuple{\Htrace,\Ttrace},j \models \varphi$ iff $\tuple{\Htrace',\Ttrace}, j \models \varphi$
\end{proposition}

\begin{definition}[$\X^{i}$]\label{def:bigX}
Let $\tuple{\Htrace,\Ttrace}$ be a \HT{-trace} of length $\lambda$ and $\rangeo{i}{0}{\lambda}$.
We denote $\X^{i}$ the trace of length $\lambda$ satisfying $X^i_k = \emptyset$ for all $\rangeo{k}{0}{i}$.
\end{definition}

\begin{lemma}\label{lem:pastocc}
    For all \HT{-traces} $\tuple{\Htrace,\Ttrace}$ of length $\lambda$, for all $\rangeo{i}{0}{\lambda}$ and for any past formula $\varphi$,
    if each present and positive occurrence of an atom from $X^i_i$
    in $\varphi$ is in the scope of negation then $\tuple{\Htrace,\Ttrace}, i \models \varphi$ iff $\tuple{\Htrace\setminus \X^i , \Ttrace},i \models \varphi$.
\end{lemma}

\begin{proofof}{Lemma \ref{lem:pastocc}}
By induction on $\varphi$.
First, note that for any formula $\phi$ of the form $\varphi\vee\psi$, $\varphi\wedge\psi$,
$\varphi\trigger\psi$ or $\varphi\since\psi$, if all present and positive occurrences of an atom $p$ are in the
scope of negation in $\phi$, then all present and positive occurrences of $p$ are also in the scope of negation
in $\varphi$ and $\psi$.

\begin{itemize}
    \item case $\bot$: clearly, $\tuple{\Htrace,\Ttrace},i\not\models\bot$  and $\tuple{\Htrace\setminus\X^i,\Ttrace},i\not\models\bot$.
    \item case $p$:
    	we consider two cases.
        If $p\not\in X^i_i$,
        $\tuple{\Htrace,\Ttrace},i\models p$ iff $p\in H_i$, iff $p\in H_i\setminus X_i$,
        iff $\tuple{\Htrace\setminus\X^i,\Ttrace},i\models p$.

        If $p\in X^i_i$, then $p$ has a present and positive occurence in $\varphi$, which is not in the scope of negation.
        Therefore, the lemma automatically holds.

    \item case $\neg \varphi$:
        $\tuple{\Htrace,\Ttrace},i \models \neg\varphi$ iff
        $\tuple{\Ttrace,\Ttrace},i \not\models \varphi$ iff
        $\tuple{\Htrace\setminus\X^i(L),\Ttrace},i \models \neg\varphi$ (because of persistency).

    \item case $\previous \varphi$:
    \begin{itemize}
        \item if $i=0$, then $\tuple{\Htrace,\Ttrace},i \not\models \previous\varphi$ and
                  $\tuple{\Htrace\setminus\X^i,\Ttrace},i \not\models \previous\varphi$.
    	\item if $i>0$, then $\tuple{\Htrace,\Ttrace},i \models \previous\varphi$ iff
    	$\tuple{\Htrace,\Ttrace},i-1 \models \varphi$.
    	%As $X^i_{i-1}=\emptyset$, we can apply the induction hypothesis, so
        Let $X^{i-1}$ be such that $X^{i-1}_{i-1}=\emptyset$. Then, we can apply the induction hypothesis, so
    	$\tuple{\Htrace,\Ttrace},i-1 \models \varphi$ iff (IH)
    	$\tuple{\Htrace\setminus\X^{i-1},\Ttrace},i-1 \models \varphi$.
    	Since $\tuple{\Htrace\setminus\X^{i-1},\Ttrace}\sim_{i} \tuple{\Htrace\setminus\X^{i},\Ttrace}$ (Proposition~\ref{prop:pure-past})
    	then $\tuple{\Htrace\setminus\X^{i-1},\Ttrace},i-1 \models \varphi$
    	iff $\tuple{\Htrace\setminus\X^{i},\Ttrace},i-1 \models \varphi$,
    	iff $\tuple{\Htrace\setminus\X^{i},\Ttrace},i \models \previous\varphi$.
    \end{itemize}

    \item case $\varphi\vee\psi$:
        $\tuple{\Htrace,\Ttrace},i \models \varphi\vee\psi$ iff
        $\tuple{\Htrace,\Ttrace},i \models \varphi$ or $\tuple{\Htrace,\Ttrace},i \models \psi$.
        Since all positive and present occurences of atoms from $X^i_i$ in $\varphi$ and $\psi$ are in the scope of
        negation, we can apply the induction hypothesis to get
        $\tuple{\Htrace\setminus\X^i,\Ttrace},i \models \varphi$ or $\tuple{\Htrace\setminus\X^i,\Ttrace},i \models \psi$.
        Therefore, $\tuple{\Htrace\setminus\X^i,\Ttrace},i \models \varphi\vee\psi$.
    \item case $\varphi\wedge\psi$: Similar as for $\varphi\vee\psi$.
    \item case $\varphi\since\psi$:
        $\tuple{\Htrace,\Ttrace},i \models \varphi\since\psi$ iff
        for some $\rangec{j}{0}{i}$, $\tuple{\Htrace,\Ttrace},j \models \psi$ and
        $\tuple{\Htrace,\Ttrace},k \models \varphi$ for all $\rangeoc{k}{j}{i}$.
        By induction we get that iff
        for some $\rangec{j}{0}{i}$, $\tuple{\Htrace\setminus \X^j,\Ttrace},j \models \psi$ and
        $\tuple{\Htrace\setminus \X^k,\Ttrace},k \models \varphi$ for all $\rangeoc{k}{j}{i}$.
        Since $\tuple{\Htrace\setminus \X^t,\Ttrace} \sim_t \tuple{\Htrace\setminus \X^i,\Ttrace}$ for all
        $\rangeco{t}{0}{i}$, by Proposition~\ref{prop:pure-past} we get that iff
        for some $\rangec{j}{0}{i}$, $\tuple{\Htrace\setminus \X^i,\Ttrace},j \models \psi$ and
        $\tuple{\Htrace\setminus \X^i,\Ttrace},k \models \varphi$ for all $\rangeoc{k}{j}{i}$.
        iff $\tuple{\Htrace\setminus\X^i,\Ttrace},i \models \varphi\since\psi$.
    \item case $\varphi\trigger\psi$:
        assume by contradiction that $\tuple{\Htrace\setminus\X^i,\Ttrace},i \not\models \varphi\trigger\psi$.
        This means that there exist $\rangec{j}{0}{i}$ such that $\tuple{\Htrace\setminus\X^i,\Ttrace},j \not\models \psi$ and
        $\tuple{\Htrace\setminus\X^i,\Ttrace},k \not\models \varphi$ for all $\rangeoc{k}{j}{i}$.
	    Since $\tuple{\Htrace\setminus \X^t,\Ttrace} \sim_t \tuple{\Htrace\setminus \X^i,\Ttrace}$ for all
	    $\rangeco{t}{0}{i}$, by Proposition~\ref{prop:pure-past} we get
		that there exist $\rangec{j}{0}{i}$ such that $\tuple{\Htrace\setminus\X^j,\Ttrace},j \not\models \psi$ and
		$\tuple{\Htrace\setminus\X^k,\Ttrace},k \not\models \varphi$ for all $\rangeoc{k}{j}{i}$.
		By induction, there exist $\rangec{j}{0}{i}$ such that $\tuple{\Htrace,\Ttrace},j \not\models \psi$ and
		$\tuple{\Htrace,\Ttrace},k \not\models \varphi$ for all $\rangeoc{k}{j}{i}$
		iff $\tuple{\Htrace,\Ttrace},i \not\models \varphi\trigger\psi$: a contradiction.
\end{itemize}
\end{proofof}

\begin{definition}
	Let $L \subseteq \A$ and let $\lambda >0$ and $\rangeco{i}{0}{\lambda}$. By $\X(L)^i$ we mean a trace of length $\lambda$ satisfying the following conditions:
		\begin{enumerate}
			\item  $L \subseteq X(L)^i_i \subseteq \A$;
			\item  $X(L)^i_t = \emptyset$ for all $\rangeco{t}{0}{i}$.
		\end{enumerate}
\end{definition}

\begin{lemma}\label{lem:support}
    Let $\tuple{\Htrace,\Ttrace}$ be a \HT{-trace} of length $\lambda$, $\varphi$ a past formula. Let us consider the set of atoms $L\subseteq\A$.
    For all $\rangeo{i}{0}{\lambda}$, if each positive occurrence of an atom from $X(L)^i_i\setminus L$ in $\varphi$ is in the scope of negation,
    $\tuple{\Htrace,\Ttrace},i\models\support{L}{\varphi}$ iff $\tuple{\Htrace\setminus\X(L)^i,\Ttrace},i\models\varphi$.
%
%     Let $\X$ be a trace of length $\lambda$ such that $X_i \supseteq L$ and
%    $X_j=\emptyset$ for $\rangeo{j}{0}{i}$.
%    If each present and positive occurrence of an atom from $X_i\setminus L$ in $\varphi$ is in the scope of negation,
%    $\tuple{\Htrace,\Ttrace},i\models\support{L}{\varphi}$ iff $\tuple{\Htrace\setminus\X,\Ttrace},i\models\varphi$.
\end{lemma}
\begin{proofof}{Lemma~\ref{lem:support}}
    By induction on $\varphi$.
    First, note that for any formula $\phi$ of the form $\varphi\vee\psi$, $\varphi\wedge\psi$,
    $\varphi\trigger\psi$ or $\varphi\since\psi$, if all present and positive occurrences of an atom $p$ are in the
    scope of negation in $\phi$, then all present and positive occurrences of $p$ are also in the scope of negation
    in $\varphi$ and $\psi$.

    \begin{itemize}
    \item case $\bot$: $\tuple{\Htrace,\Ttrace},i\not\models\bot$ and $\tuple{\Htrace\setminus\X(L)^i,\Ttrace},i\not\models\bot$.
    \item case $p\not\in L$: we consider the following two cases
    	\begin{itemize}
    		\item If $p\not\in L$, $\support{L}{p}=p$ and, by definition, $p \not \in X(L)^i_i$.
    		      Therefore,  $\tuple{\Htrace,\Ttrace},i\models \support{L}{p}$ iff $\tuple{\Htrace,\Ttrace},i\models p $, iff $p\in H_i$
    		      iff $p\in H_i\setminus X(L)_i^i$, iff $\tuple{\Htrace\setminus\X(L)^i,\Ttrace},i\models p$.

    		\item If $p\in L$ then $p \not \in X(L)^i_i \setminus L$ and $\support{L}{p}=\bot$.
    		Therefore, we get that $\tuple{\Htrace,\Ttrace},i\not\models \support{L}{p}$ and $\tuple{\Htrace\setminus\X^i(L),\Ttrace},i\not\models p$.
    	\end{itemize}

    \item case $\neg\varphi$:
        $\tuple{\Htrace,\Ttrace},i\models \support{L}{\neg\varphi}$ iff $\tuple{\Htrace,\Ttrace},i\models \neg\varphi$,
        iff $\tuple{\Ttrace,\Ttrace},i\not\models \varphi$, iff $\tuple{\Htrace\setminus\X^i(L),\Ttrace},i\models \neg\varphi$.
    \item case $\previous\varphi$:
        $\tuple{\Htrace,\Ttrace},i\models\support{L}{\previous\varphi}$ iff $\tuple{\Htrace,\Ttrace},i\models\previous\varphi$.
        \begin{itemize}
        	\item If $i=0$, then both $\tuple{\Htrace,\Ttrace},0\not\models\previous\varphi$ and
        	      $\tuple{\Htrace\setminus\X(L)^0,\Ttrace},0\not\models\previous\varphi$.
        	\item If $i>0$, then
        	        $\tuple{\Htrace,\Ttrace},i\models\previous\varphi$ iff $\tuple{\Htrace,\Ttrace},i-1\models\varphi$.
        	        By definition, $X(L)^i_{j} = \emptyset$
        	        for $\rangeo{j}{0}{i}$ so by Lemma~\ref{lem:pastocc},
        	        $\tuple{\Htrace\setminus\X^{i}(L),\Ttrace},i-1\models\varphi$,
        	        iff $\tuple{\Htrace\setminus\X^i(L),\Ttrace},i\models\previous\varphi$.

        \end{itemize}

    \item case $\varphi\wedge\psi$:
        $\tuple{\Htrace,\Ttrace},i\models\support{L}{\varphi\wedge\psi}$\\
        iff $\tuple{\Htrace,\Ttrace},i\models\support{L}{\varphi}$ and $\tuple{\Htrace,\Ttrace},i\models\support{L}{\psi}$,\\
        iff (IH) $\tuple{\Htrace\setminus\X^i(L),\Ttrace},i\models\varphi$ and $\tuple{\Htrace\setminus\X^i(X),\Ttrace},i\models\psi$,\\
        iff $\tuple{\Htrace\setminus\X^i(L),\Ttrace},i\models\varphi\wedge\psi$.
    \item case $\varphi\vee\psi$:
        $\tuple{\Htrace,\Ttrace},i\models\support{L}{\varphi\vee\psi}$\\
        iff $\tuple{\Htrace,\Ttrace},i\models\support{L}{\varphi}$ or $\tuple{\Htrace,\Ttrace},i\models\support{L}{\psi}$,\\
        iff (IH) $\tuple{\Htrace\setminus\X^i(L),\Ttrace},i\models\varphi$ or $\tuple{\Htrace\setminus\X^i(L),\Ttrace},i\models\psi$,\\
        iff $\tuple{\Htrace\setminus\X^i(L),\Ttrace},i\models\varphi\vee\psi$.
%    \item case $\varphi\trigger\psi$
%        $\tuple{\Htrace,\Ttrace},i\models\support{L}{\varphi\trigger\psi}$\\
%        iff
%        \begin{enumerate}
%        	\item
%        \end{enumerate}
%
%
%        $\tuple{\Htrace,\Ttrace},i\models\support{L}{\psi}$ and, $\tuple{\Htrace,\Ttrace},i\models\support{L}{\varphi}$
%            or $\tuple{\Htrace,\Ttrace},i\models\previous(\varphi\trigger\psi)$,\\
%        iff (IH) $\tuple{\Htrace\setminus\X,\Ttrace},i\models\psi$  and, $\tuple{\Htrace\setminus\X,\Ttrace},i\models\varphi$
%                 or $\tuple{\Htrace,\Ttrace},i\models\previous(\varphi\trigger\psi)$,\\
%        iff $\tuple{\Htrace\setminus\X,\Ttrace},i\models\psi$  and, $\tuple{\Htrace\setminus\X,\Ttrace},i\models\varphi$
%            or $\tuple{\Htrace,\Ttrace},i-1\models\varphi\since\psi$,\\
%            iff, by Lemma~\ref{lem:pastocc} as $X_j=\emptyset$ for $\rangeo{j}{0}{i}$,
%            $\tuple{\Htrace\setminus\X,\Ttrace},i\models\psi$  and, $\tuple{\Htrace\setminus\X,\Ttrace},i\models\varphi$
%            or $\tuple{\Htrace\setminus\X,\Ttrace},i-1\models\varphi\since\psi$,\\
%        iff $\tuple{\Htrace\setminus\X,\Ttrace},i\models\psi$  and, $\tuple{\Htrace\setminus\X,\Ttrace},i\models\varphi$
%            or $\tuple{\Htrace\setminus\X,\Ttrace},i\models\previous(\varphi\since\psi)$,\\
%        iff $\tuple{\Htrace\setminus\X,\Ttrace},i\models\varphi\since\psi$.
    \item case $\varphi\since\psi$:
        $\tuple{\Htrace,\Ttrace},i\models\support{L}{\varphi\since\psi}$ iff
        \begin{enumerate}
        	\item $\tuple{\Htrace,\Ttrace},i\models\support{L}{\psi}$ iff  $\tuple{\Htrace\setminus\X^i(L),\Ttrace},i\models\psi$ (IH) or
        	\item $\tuple{\Htrace,\Ttrace},i\models\support{L}{\varphi}$ and $\tuple{\Htrace,\Ttrace},i\models\previous(\varphi\since\psi)$
        	iff  $\tuple{\Htrace\setminus\X^i(L),\Ttrace},i\models\varphi$ (IH) and $\tuple{\Htrace,\Ttrace},i\models\previous(\varphi\since\psi)$
        	iff  $\tuple{\Htrace\setminus\X^i(L),\Ttrace},i\models\varphi$ and $\tuple{\Htrace\setminus \X^i(L),\Ttrace},i\models\previous(\varphi\since\psi)$
        \end{enumerate}
        From the previous items we conclude iff
        $\tuple{\Htrace\setminus\X^i(L),\Ttrace},i\models \psi \vee \left( \varphi \wedge \previous \left(\varphi \since \psi\right)\right)$
        iff $\tuple{\Htrace\setminus\X^i(L),\Ttrace},i\models \varphi \since \psi$.

    \item case $\varphi\trigger\psi$:
		$\tuple{\Htrace,\Ttrace},i\not \models\support{L}{\varphi\trigger\psi}$ iff
		\begin{enumerate}
			\item $\tuple{\Htrace,\Ttrace},i\not \models\support{L}{\psi}$ iff  $\tuple{\Htrace\setminus\X^i(L),\Ttrace},i\not \models\psi$ (IH) or
			\item $\tuple{\Htrace,\Ttrace},i\not \models\support{L}{\varphi}$ and $\tuple{\Htrace,\Ttrace},i\not \models\previous(\varphi\trigger\psi)$
			iff  $\tuple{\Htrace\setminus\X^i(L),\Ttrace},i\not \models\varphi$ (IH) and $\tuple{\Htrace,\Ttrace},i\not \models\previous(\varphi\trigger\psi)$
			iff  $\tuple{\Htrace\setminus\X^i(L),\Ttrace},i\not \models\varphi$ and $\tuple{\Htrace\setminus \X^i(L),\Ttrace},i\not \models\previous(\varphi\trigger\psi)$
\end{enumerate}
From the previous items we conclude iff
$\tuple{\Htrace\setminus\X^i(L),\Ttrace},i\not \models \psi \wedge \left( \varphi \vee \previous \left(\varphi \trigger \psi\right)\right)$
iff $\tuple{\Htrace\setminus\X^i(L),\Ttrace},i\not \models \varphi \trigger \psi$.

%        iff $\tuple{\Htrace,\Ttrace},i\models\support{L}{\psi}$ or, \\
%        iff (IH) $\tuple{\Htrace\setminus\X,\Ttrace},i\models\psi$  or, $\tuple{\Htrace\setminus\X,\Ttrace},i\models\varphi$
%                 and $\tuple{\Htrace,\Ttrace},i\models\previous(\varphi\since\psi)$,\\
%        iff $\tuple{\Htrace\setminus\X,\Ttrace},i\models\psi$  or, $\tuple{\Htrace\setminus\X,\Ttrace},i\models\varphi$
%            and $\tuple{\Htrace,\Ttrace},i-1\models\varphi\since\psi$,\\
%        iff,  by Lemma~\ref{lem:pastocc} as $X_j=\emptyset$ for $\rangeo{j}{0}{i}$,
%        $\tuple{\Htrace\setminus\X,\Ttrace},i\models\psi$  or, $\tuple{\Htrace\setminus\X,\Ttrace},i\models\varphi$
%            and $\tuple{\Htrace\setminus\X,\Ttrace},i-1\models\varphi\since\psi$,\\
%        iff $\tuple{\Htrace\setminus\X,\Ttrace},i\models\psi$  or, $\tuple{\Htrace\setminus\X,\Ttrace},i\models\varphi$
%            and $\tuple{\Htrace\setminus\X,\Ttrace},i\models\previous(\varphi\since\psi)$,\\
%        iff $\tuple{\Htrace\setminus\X,\Ttrace},i\models\varphi\since\psi$.
    \end{itemize}
\end{proofof}

\begin{proofof}{Theorem \ref{thm:tcompletion}}
From left to right, let us assume towards a contradiction that \Ttrace\ is a temporal answer set of $P$, 
but \Ttrace\ is not an \LTLf-model of $\tCF{P}$.
By construction, if $\Ttrace$ is a temporal answer set of $P$ then $\T$ is an \LTLf{} model of $P$ so $\T, 0 \models P$.
Therefore, $\T,0 \models r$, for all $r \in P$ such that $\Head{r} = \bot$ and
$\T, 0 \models r$ for all $r \in F(P)$.
Since $\T, 0 \not \models \tCF{P}$, there exists $a \in \A$ such that 

\begin{align*}
	\T, 0 \not \models \alwaysF\big( a \leftrightarrow
	\ \bigvee_{r\in\initial{P}, a\in\Head{r}} (\initially\wedge S(r,a))
	\vee \bigvee_{r\in\dynamic{P}, a\in\Head{r}} (\neg\initially\wedge S(r,a))\big)
\end{align*}

So, there exists $\rangeco{i}{0}{\lambda}$ such that

\begin{align*}
	\T, i \not \models a \leftrightarrow
	\ \bigvee_{r\in\initial{P}, a\in\Head{r}} (\initially\wedge S(r,a))
	\vee \bigvee_{r\in\dynamic{P}, a\in\Head{r}} (\neg\initially\wedge S(r,a))
\end{align*}

We consider two cases: 

\begin{enumerate}
	\item $\T, i \models a$ and $\T, i \not \models 
	\ \bigvee_{r\in\initial{P}, a\in\Head{r}} (\initially\wedge S(r,a))
	\vee \bigvee_{r\in\dynamic{P}, a\in\Head{r}} (\neg\initially\wedge S(r,a))$: 
	
	\begin{itemize}
		\item If $i = 0$ then we get that for all $r \in \initial{P}$, if $a \in \Head{r}$ then $\tuple{\Ttrace,\Ttrace}, 0 \not \models S(r,a)$. 
		Therefore, for all $r \in \initial{P}$, if $a \in \Head{r}$ then $\tuple{\Ttrace,\Ttrace}, 0 \not \models \S(r,a)$.
		Let \Htrace\ be a trace of length $\lambda$ such that $H_0=T_0\setminus\{a\}$ and $H_i=T_i$ for $\rangeo{i}{1}{\lambda}$.
		Clearly, $\Htrace < \Ttrace$.
		We show a contradiction by proving that $\tuple{\Htrace,\Ttrace}\models P$:
		\begin{enumerate}
			\item  $\tuple{\Htrace,\Ttrace},0\models \initial{P}$: note that $\tuple{\Ttrace,\Ttrace},0\models \body{r}\to\Head{r}$ for all $r\in\initial{P}$, iff for any $r\in\initial{P}$, $\tuple{\Ttrace,\Ttrace},0\not\models \body{r}$ or $\tuple{\Ttrace,\Ttrace},0\models \Head{r}$.
			If $\tuple{\Ttrace,\Ttrace},0\not\models \body{r}$ then, by persistence, $\tuple{\Htrace,\Ttrace},0\not\models \body{r}$, and $\tuple{\Htrace,\Ttrace},0\models \body{r}\to\Head{r}$.
			If $\tuple{\Ttrace,\Ttrace},0\models \body{r}$, then $\tuple{\Ttrace,\Ttrace},0\models \Head{r}$.
			There are two cases.
			\begin{itemize}
				\item Case $a\not\in\Head{r}$:
				$\tuple{\Ttrace,\Ttrace},0\models \Head{r}$ so there is some $p\in\Head{r}$ such that $p\in T_0$ and $p \not = a$.  Then, $p\in H_0$, $\tuple{\Htrace,\Ttrace},0\models \Head{r}$ and $\tuple{\Htrace,\Ttrace},0\models r$.
				\item Case $a\in\Head{r}$:
				We know that $\tuple{\Ttrace,\Ttrace},0\not\models \body{r}\wedge \bigwedge_{p\in \Head{r}\setminus \{a\}} \neg p$.
				Since, by assumption, $\tuple{\Ttrace,\Ttrace},0\models \body{r}$, it follows 
				$\tuple{\Ttrace,\Ttrace}, 0\not\models  \bigwedge_{p\in \Head{r}\setminus \{a\}} \neg p$
				Therefore, there is $p\in\Head{r}\setminus\{a\}$ such that $p\in T_0$. Then $p\in H_0$, $\tuple{\Htrace,\Ttrace},0\models\Head{r}$ and $\tuple{\Htrace,\Ttrace},0\models r$.
%				We have already shown that, if $\tuple{\Ttrace,\Ttrace},0\not\models \body{r}$,
%				then $\tuple{\Htrace,\Ttrace},0\models r$.
%				If , there is
%				some 
			\end{itemize}
				As $r$ is chosen arbitrarily, $\tuple{\Htrace,\Ttrace}\models \initial{P}$.

			\item $\tuple{\Htrace,\Ttrace},0\models \dynamic{P}$: $\tuple{\Ttrace,\Ttrace}\models \dynamic{P}$, then $\tuple{\Ttrace,\Ttrace},i\models \body{r}\to\Head{r}$
			for all $r\in\dynamic{P}$ and $\rangeo{i}{1}{\lambda}$. Then, for any $r\in\dynamic{P}$ and
			$\rangeo{i}{1}{\lambda}$, $\tuple{\Ttrace,\Ttrace},i\not\models \body{r}$ or $\tuple{\Ttrace,\Ttrace},i\models \Head{r}$.
			If $\tuple{\Ttrace,\Ttrace},i\not\models \body{r}$ then, by persistence, $\tuple{\Htrace,\Ttrace},i\not\models \body{r}$,
			and $\tuple{\Htrace,\Ttrace},i\models r$.
			If, $\tuple{\Ttrace,\Ttrace},i\models \Head{r}$, there is some $p\in\Head{r}$ such that $p\in T_i$.
			$H_i = T_i$, so $p\in H_i$ and $\tuple{\Htrace,\Ttrace},i\models \Head{r}$.
			Then $\tuple{\Htrace,\Ttrace},i\models r$. As $r$ and $i$ are chosen arbitrarily,
			$\tuple{\Htrace,\Ttrace}\models \dynamic{P}$.
			\item $\tuple{\Htrace,\Ttrace},0\models \final{P}$: final rules are constraints, so $\tuple{\Ttrace,\Ttrace}\models \final{P}$ implies $\tuple{\Htrace,\Ttrace}\models \final{P}$.
		\end{enumerate}
		We showed that $\tuple{\Htrace, \Ttrace}, 0 \models P$ : a contradiction.

		\item If $i > 0$: we follow a very similar reasoning as for the case $i=0$.
		
	\end{itemize}	
	
%	
%	
%	in this case we conclude that for all $r \in P$ one of the following items hold
%		\begin{itemize}
%			\item if $r \in \initial{P}$ and  $a \in \Head{r}$ then either $\T, 0 \not \models \body{r}$ or $\T, 0 \models \Head{r} \setminus \lbrace a \rbrace$ 
%			\item if $r \in \dynamic{P}$ and  $a \in \Head{r}$ then either $\T, i \not \models \body{r}$ or $\T, i \models \Head{r} \setminus \lbrace a \rbrace$, for some $i > 0$.\footnote{\color{red}M: finish this part}
%		\end{itemize}	
		
	\item $\T, i \not \models a$ but $\T, i \models
	\ \bigvee_{r\in\initial{P}, a\in\Head{r}} (\initially\wedge S(r,a))
	\vee \bigvee_{r\in\dynamic{P}, a\in\Head{r}} (\neg\initially\wedge S(r,a))$: again, we consider two cases here
		\begin{itemize}
			\item there exists $r\in\initial{P}, a\in\Head{r}$ and  $\T, i \models \initially\wedge S(r,a)$: in this case, it follows that $i= 0$, so $\T, 0 \models a$ and $\T, 0 \models  S(r,a)$.
			Therefore, $\T, 0 \models \body{r}$.
			Since $\T, 0 \models r$ and $\T, 0 \models \body{r}$ then $\T, 0 \models p$ for some $p \in \Head{r}$, which contradicts 
			$\T, 0 \not \models a$ and $\T, 0 \models \neg q$ for all $p \in \Head{r} \setminus\lbrace a\rbrace$.
			
			\item there exists $r\in\dynamic{P}, a\in\Head{r}$ and  $\T, i \models  \neg \initially\wedge S(r,a)$: in this case we conclude that $i >0$ and so $\T, i \models a$ and $\T, i \models  S(r,a)$.
			Therefore, $\T, i \models \body{r}$.
			Since $\T, 0 \models r$ and $\T, i \models \body{r}$ then $\T, i \models q$ for some $q \in \Head{r}$.
			However, from $\T, i \models  S(r,a)$ and $\T, i \not \models a$ we conclude that $\T, i \not \models p$ for all $p \in \Head{r}$: a contradiction.
		\end{itemize}
		
\end{enumerate}

For the converse direction, assume, again, by contradiction that $\tuple{\Ttrace,\Ttrace}$ is not a \TELf{} model of $P$.
We consider two cases:

\begin{enumerate}
	\item $\tuple{\Ttrace,\Ttrace},0 \not \models P$.
	Therefore, there exists $r \in P$ such that $\tuple{\Ttrace,\Ttrace}, 0 \not \models r$.
	Clearly, $r$ cannot be a constraint, otherwise we would already reach a contradiction.
	We still have to check two cases:
	\begin{itemize}
		\item If $r \in \initial{P}$, then $\tuple{\Ttrace,\Ttrace}, 0 \models \body{r}$ and $\tuple{\Ttrace,\Ttrace},0 \not \models \Head{r}$.
		Take any $a \in \Head{r}$. It follows that $\tuple{\Ttrace,\Ttrace}, 0 \models \initially \wedge S(r,a)$.
		so $\tuple{\Ttrace,\Ttrace}, 0 \not \models \tCFa{P}{a}$: a contradiction.
		\item If $r \in \dynamic{P}$ we follow a similar reasoning as for the previous case.
	\end{itemize}	

	\item $\tuple{\Ttrace,\Ttrace},0 \models P$ but $\tuple{\Htrace,\Ttrace}, 0 \models P$ for some $\Htrace < \Ttrace$.
	By definition, there exists $i \ge 0$ such that $H_i \subset T_i$. Let us take the smallest $i$ satisfying this property.
	Therefore, $H_j = T_j$ for all $\rangeco{j}{0}{i}$. 
	.Moreover, Let us take $a \in T_i\setminus H_i$ and let us proceed depending on the value of $i$
	
	\begin{itemize}
		\item If $i > 0 $ and $a \in T_i$ then $\tuple{\Ttrace,\Ttrace},i \models \tCFa{P}{a}$ then there exists $r \in \dynamic{P}$ such that $\tuple{\Ttrace,\Ttrace},i \models S(r,a)$.
			  Therefore $\Head{t} \setminus \lbrace a \rbrace \cup T_i = 0$
			  Since $a \not \in H_i$ then $\tuple{\Htrace,\Ttrace}, i \not \models \Head{r}$ and $\tuple{\Htrace,\Ttrace}, i \not \models \body{r}$.

			  At this point of the proof we have that $\tuple{\Ttrace,\Ttrace},i \models \body{r}$, $\tuple{\Htrace,\Ttrace},i \not\models \body{r}$ and $T_j\setminus H_j = \emptyset$ for any $j<i$.
	%\footnote{\color{red}M: check this paragraph}	
	By Lemma~\ref{lem:pastocc}, there must be some $b\in T_i\setminus H_i$ with a present and positive occurence
		in $\body{r}$ that is not in the scope of negation.
		Then, for any $a\in T_i\setminus H_i$, there is some $b\in T_i\setminus H_i$ such that $(a,b)\in G(\dynamic{P})$.
		$P$ is tight, so $G(\dynamic{P})$ is acyclic. Then, there is a topological ordering of the nodes in $G(\dynamic{P})$,
		and therefore of the atoms in $T_i\setminus H_i$, such that if $a,b \in T_i\setminus H_i$ and $(a,b)\in G(\dynamic{P})$,
		then $a$ appears before $b$ in the topoligical ordering.
		Then, there is no outgoing edge from the last node in the ordering, which contradict the fact that,
		for any $a\in T_i\setminus H_i$, there is some $b\in T_i\setminus H_i$ such that $(a,b)\in G(\dynamic{P})$.\\
		
		\item If $i = 0$ we proceed as in the previous case.
	\end{itemize}

\end{enumerate}

\end{proofof}

\begin{proofof}{Theorem \ref{thm:loops}}

We first prove that if \Ttrace\ is a temporal answer set of $P$, then \Ttrace\ is a \LTLf-model of \tCF{P} and \LF{P}.
The proof for \tCF{P} is the same as for Theorem~\ref{thm:tcompletion}.
Remains to prove that \Ttrace\ is a \LTLf-model of \LF{P}.
Assume by contradiction that $\tuple{\Ttrace,\Ttrace}\not\models \LF{P}$. Two different cases must be considered:
\begin{itemize}
    \item
        there is a loop $L$ in \Graph{\dynamic{P}} such that
        $\tuple{\Ttrace,\Ttrace},i\not\models \bigvee_{a\in L} a \rightarrow \mathit{ES}_{\dynamic{P}}(L)$
        for some $\rangeo{i}{1}{\lambda}$, or
    \item
        there is a loop $L$ in \Graph{\initial{P}} such that
        $\tuple{\Ttrace,\Ttrace},0\not\models \bigvee_{a\in L} a \rightarrow \mathit{ES}_{\initial{P}}(L)$.
\end{itemize}

For the first case, let \Htrace\ be a trace of length $\lambda$ such that $H_i=T_i\setminus L$ and
$H_k=T_k$ otherwise.
We show that $\tuple{\Htrace,\Ttrace},0\models P$, which will contradict the hypothesis \Ttrace\ is a \TELf-model of $P$:
\begin{enumerate}
    \item
        $\tuple{\Htrace,\Ttrace},0\models \initial{P}$: follows from  $\tuple{\Ttrace,\Ttrace}\models \initial{P}$
        by Lemma~\ref{lem:pastocc} as $T_0\setminus H_0 = \emptyset$.
    \item
        $\tuple{\Htrace,\Ttrace},0\models \final{P}$: follows from  $\tuple{\Ttrace,\Ttrace}\models \final{P}$ as rules in \final{P} are
        constraints.
   \item
        $\tuple{\Htrace,\Ttrace},0\models \dynamic{P}$: $\tuple{\Ttrace,\Ttrace},0\models \dynamic{P}$ since $\Ttrace$ is a \TELf{}-model of $P$.
        Therefore, $\tuple{\Ttrace,\Ttrace},k \models r$ for all $\rangeo{k}{1}{\lambda}$ and for all $r\in\dynamic{P}$.
        Then, $\tuple{\Ttrace,\Ttrace},k \not\models \body{r}$ or $\tuple{\Ttrace,\Ttrace},k \models \Head{r}$.
        If $\tuple{\Ttrace,\Ttrace},k \not\models \body{r}$, by persistence, $\tuple{\Htrace,\Ttrace},k \not\models \body{r}$ and
        $\tuple{\Htrace,\Ttrace},k \models r$.
        If $\tuple{\Ttrace,\Ttrace},k \models \body{r}$, then $\tuple{\Ttrace,\Ttrace},k \models \Head{r}$.

        In the case $k\neq i$, $H_k=T_k$ and $\tuple{\Ttrace,\Ttrace}, k \models \Head{r}$ imply  $\tuple{\Htrace,\Ttrace},k \models \Head{r}$.% follows
        %from $\tuple{\Ttrace,\Ttrace},k \models \Head{r}$,
        %and then $\tuple{\Ttrace,\Ttrace},k \models r$.
%
        In the case $k=i$, we have two cases.
        \begin{itemize}
            \item
                if $\tuple{\Ttrace,\Ttrace},i\not\models\support{L}{\body{r}}$, then,
                as $(T_i\setminus H_i)\setminus L=\emptyset$ and $T_k\setminus H_k=\emptyset$ for $k<i$,
                by Lemma~\ref{lem:support},
                $\tuple{\Htrace,\Ttrace},i \not\models \body{r}$. So $\tuple{\Htrace,\Ttrace},i \models r$.
            \item if $\tuple{\Ttrace,\Ttrace},i\models\support{L}{\body{r}}$ and $\Head{r}\cap L =\emptyset$ then 
            $\tuple{\Htrace,\Ttrace},i \models \Head{r}$ follows from $\tuple{\Ttrace,\Ttrace},i \models \Head{r}$ and $\tuple{\Htrace,\Ttrace},i \models r$.
            \item if $\tuple{\Ttrace,\Ttrace},i\models\support{L}{\body{r}}$ and $\Head{r}\cap L =\emptyset$ then
                        \begin{itemize}
                            \item if there is some $p\in\Head{r}\setminus L$ such that $p\in T_i$, then $p\in H_i$. So
                                $\tuple{\Htrace,\Ttrace},i \models \Head{r}$ and then $\tuple{\Htrace,\Ttrace},i \models r$.
                            \item if there is no $p\in\Head{r}\setminus L$ such that $p\in T_i$, then
                                $\tuple{\Ttrace,\Ttrace},i\models \bigwedge_{p\in \Head{r}\setminus L} \neg p$.
                                As we also have $\tuple{\Ttrace,\Ttrace},i\models\support{L}{\body{r}}$,
                                $\tuple{\Ttrace,\Ttrace},i\models \bigvee_{a\in L} a \rightarrow \mathit{ES}_{\dynamic{P}}(L)$,
                                which contradict our hypothesis.
                        \end{itemize}
        \end{itemize}
\end{enumerate}

The proof of the second case follows a similar reasoning as for the first one.

%
%%-------------------
Next, we prove that if \Ttrace\ is a \LTLf-model of \tCF{P} and \LF{P}, then \Ttrace\ is a \TELf-model of $P$.
The proof for $\tuple{\Ttrace,\Ttrace}\models P$ is the same as for Theorem~\ref{thm:tcompletion}.
Remains to prove that there is no $\Htrace<\Ttrace$ such that $\tuple{\Htrace,\Ttrace}\models P$.

Let assume that there exists such a trace $\Htrace$, and let $i$ be the smallest time point such that $H_i\subset T_i$.
Therefore, $H_k = T_k$ for all $\rangeco{k}{0}{i}$.

\begin{itemize}
	\item If $i>0$:
	Let $a\in T_i\setminus H_i$. $\tuple{\Ttrace,\Ttrace}\models \tCF{P}$, so $\tuple{\Ttrace,\Ttrace},i \models a \leftrightarrow
	\bigvee_{r\in\dynamic{P}, a\in\Head{r}}(\body{r}\wedge\bigwedge_{p\in \Head{r}\setminus \{a\}} \neg p)$.
	As $a\in T_i$, there is some rule $r\in\dynamic{P}$ such that $a\in\Head{r}$,
	$\tuple{\Ttrace,\Ttrace},i \models \body{r}$, and $\tuple{\Ttrace,\Ttrace},i \models \bigwedge_{p\in \Head{r}\setminus \{a\}} \neg p$.
	$\tuple{\Htrace,\Ttrace}\models P$, so $\tuple{\Htrace,\Ttrace},i \models \body{r} \rightarrow a \vee\bigvee_{p\in\Head{r}\setminus\{a\}} p$.
	Then, $\tuple{\Htrace,\Ttrace},i \not\models \body{r}$ or $\tuple{\Htrace,\Ttrace},i \models a$ or
	$\tuple{\Htrace,\Ttrace},i \models \bigvee_{p\in\Head{r}\setminus\{a\}} p$.
	As $a\not\in H_i$, $\tuple{\Htrace,\Ttrace},i \not\models a$.
	As $\tuple{\Ttrace,\Ttrace},i \models \bigwedge_{p\in \Head{r}\setminus \{a\}} \neg p$,
	$\tuple{\Htrace,\Ttrace},i \not\models \bigvee_{p\in\Head{r}\setminus\{a\}} p$.
	So, $\tuple{\Htrace,\Ttrace},i \not\models \body{r}$.

$\tuple{\Ttrace,\Ttrace},i \models \body{r}$, $\tuple{\Htrace,\Ttrace},i \not\models \body{r}$ and $T_j\setminus H_j = \emptyset$ for $j<i$, so,
by Lemma~\ref{lem:pastocc}, there must be some $b\in T_i\setminus H_i$ with a present and positive occurence in $\body{r}$ that is not in the scope of negation.
Therefore, for any $a\in T_i\setminus H_i$, there is some $b\in T_i\setminus H_i$ such that $(a,b)\in G(\dynamic{P})$.
It implies a loop $L$ in $\dynamic{P}$, with $L\subseteq T_i\setminus H_i$.

The strongly connected components (SCC) of the dependency graph of $\dynamic{P}$ over $T_i\setminus H_i$ form
a directed acyclic graph, so there is some SCC $L$, such that,
for any $a\in L$, there is no $b\in (T_i\setminus H_i)\setminus L$ such that $(a,b)\in \Graph{\dynamic{P}}$.

For any $a\in T_i\setminus H_i$, $\tuple{\Htrace,\Ttrace},i\not\models \body{r}$, for all $r \in \dynamic{P}$
such that $a\in\Head{r}$ and $\tuple{\Ttrace,\Ttrace},i \models \bigwedge_{p\in \Head{r}\setminus \{a\}} \neg p$.
So $\tuple{\Htrace,\Ttrace},i\not\models \body{r}$, for all $r \in \dynamic{P}$ such that $L\cap\Head{r}\neq \emptyset$
and $\tuple{\Ttrace,\Ttrace},i \models \bigwedge_{p\in \Head{r}\setminus L} \neg p$.
Let \X\ be a trace of length $\lambda$ with $X_i=L$ and $X_j=\emptyset$ for $ j\neq i$.
For any $a\in L$ there is no $b\in (T_i\setminus H_i)\setminus L$ such that $(a,b)\in \Graph{\dynamic{P}}$,
so all positive and present occurences of atoms from $L$ in $\body{r}$ are in the scope of negation.
Then, we can apply Lemma~\ref{lem:pastocc}, and get that
$\tuple{\Ttrace\setminus \X,\Ttrace},i\not\models \body{r}$, for all $r \in \dynamic{P}$ such that $L\cap\Head{r}\neq \emptyset$
and $\tuple{\Ttrace,\Ttrace},i \models \bigwedge_{p\in \Head{r}\setminus L} \neg p$.
Then, as $X_i\setminus L = \emptyset$, by Lemma~\ref{lem:support},
$\tuple{\Ttrace,\Ttrace},i\not\models \support{L}{\body{r}}$, for all $r \in \dynamic{P}$ such that $L\cap\Head{r}\neq \emptyset$
and $\tuple{\Ttrace,\Ttrace},i \models \bigwedge_{p\in \Head{r}\setminus L} \neg p$.
So, $\tuple{\Ttrace,\Ttrace},i\not\models \bigvee_{a\in L} a \rightarrow \mathit{ES}_{\dynamic{P}}(L)$,
and then $\tuple{\Ttrace,\Ttrace}\not\models \LF{P}$. Contradiction.\\

\item Case $i=0$: we reach a contradiction in a similar way as above.
	
\end{itemize}
\end{proofof}

\begin{proofof}{Theorem \ref{thm:loops0}}

We first prove that if \Ttrace\ is a temporal answer set of $P$, then \Ttrace\ is a \LTLf-model of $P \cup \LF{P}$.
\Ttrace\ is a temporal answer set of $P$, so \Ttrace\ is a \LTLf-model of $P$.
We can show that \Ttrace\ is a \LTLf-model of \LF{P} the same way as for Theorem~\ref{thm:loops}.\\

%-------------------
Next, we prove that if \Ttrace\ is a \LTLf-model of $P \cup \LF{P}$, then \Ttrace\ is a temporal answer set of $P$.
It amounts to showing that there is no $\Htrace<\Ttrace$ such that $\tuple{\Htrace,\Ttrace}\models P$.
Let assume that there is such a trace $\Htrace$, and let $i$ be the smallest time point such that $H_i\subset T_i$.
\begin{enumerate}
	\item If $i>0$,
	Let $a\in T_i\setminus H_i$. $\tuple{\Ttrace,\Ttrace}\models \LF{P}$, so $\tuple{\Ttrace,\Ttrace},i \models a \leftrightarrow
	\bigvee_{r\in\dynamic{P}, a\in\Head{r}}(\support{a}{\body{r}}\wedge\bigwedge_{p\in \Head{r}\setminus \{a\}} \neg p)$.
	As $a\in T_i$, there exists $r\in\dynamic{P}$ such that $a\in\Head{r}$,
	$\tuple{\Ttrace,\Ttrace},i \models \support{a}{\body{r}}$ and $\tuple{\Ttrace,\Ttrace},i \models \bigwedge_{p\in \Head{r}\setminus \{a\}} \neg p$.
	$\tuple{\Htrace,\Ttrace}\models P$, so $\tuple{\Htrace,\Ttrace},i \models \body{r} \rightarrow a \vee\bigvee_{p\in\Head{r}\setminus\{a\}} p$.
	Then, $\tuple{\Htrace,\Ttrace},i \not\models \body{r}$ or $\tuple{\Htrace,\Ttrace},i \models a$ or
	$\tuple{\Htrace,\Ttrace},i \models \bigvee_{p\in\Head{r}\setminus\{a\}} p$.
	As $a\not\in H_i$, $\tuple{\Htrace,\Ttrace},i \not\models a$.
	As $\tuple{\Ttrace,\Ttrace},i \models \bigwedge_{p\in \Head{r}\setminus \{a\}} \neg p$,
	$\tuple{\Htrace,\Ttrace},i \not\models \bigvee_{p\in\Head{r}\setminus\{a\}} p$.
	So, $\tuple{\Htrace,\Ttrace},i \not\models \body{r}$.

	$\tuple{\Ttrace,\Ttrace},i \models \support{a}{\body{r}}$, $\tuple{\Htrace,\Ttrace},i \not\models \body{r}$ and $T_j\setminus H_j = \emptyset$ for $j<i$, so,
	by Lemma~\ref{lem:support}, there must be some $b\in T_i\setminus H_i$ with a present and positive occurence in $\body{r}$ that is not in the scope of negation.
	Therefore, for any $a\in T_i\setminus H_i$, there is some $b\in T_i\setminus H_i$ such that $(a,b)\in G(\dynamic{P})$.
	It implies a loop $L$ in $\dynamic{P}$, with $L\subseteq T_i\setminus H_i$.
	
	The strongly connected components (SCC) of the dependency graph of $\dynamic{P}$ over $T_i\setminus H_i$ form
	a directed acyclic graph, so there is some SCC $L$, such that,
	for any $a\in L$, there is no $b\in (T_i\setminus H_i)\setminus L$ such that $(a,b)\in \Graph{\dynamic{P}}$.

	For any $a\in T_i\setminus H_i$, $\tuple{\Htrace,\Ttrace},i\not\models \body{r}$, for all $r \in \dynamic{P}$
	such that $a\in\Head{r}$ and $\tuple{\Ttrace,\Ttrace},i \models \bigwedge_{p\in \Head{r}\setminus \{a\}} \neg p$.
	So $\tuple{\Htrace,\Ttrace},i\not\models \body{r}$, for all $r \in \dynamic{P}$ such that $L\cap\Head{r}\neq \emptyset$
	and $\tuple{\Ttrace,\Ttrace},i \models \bigwedge_{p\in \Head{r}\setminus L} \neg p$.
	Let \X\ be a trace of length $\lambda$ with $X_i=L$ and $X_j=\emptyset$ for $ j\neq i$.
	For any $a\in L$ there is no $b\in (T_i\setminus H_i)\setminus L$ such that $(a,b)\in \Graph{\dynamic{P}}$,
	so all positive and present occurences of atoms from $L$ in $\body{r}$ are in the scope of negation.
	Then, we can apply Lemma~\ref{lem:pastocc}, and get that
	$\tuple{\Ttrace\setminus \X,\Ttrace},i\not\models \body{r}$, for all $r \in \dynamic{P}$ such that $L\cap\Head{r}\neq \emptyset$
	and $\tuple{\Ttrace,\Ttrace},i \models \bigwedge_{p\in \Head{r}\setminus L} \neg p$.
	Then, as $X_i\setminus L = \emptyset$, by Lemma~\ref{lem:support},
	$\tuple{\Ttrace,\Ttrace},i\not\models \support{L}{\body{r}}$, for all $r \in \dynamic{P}$ such that $L\cap\Head{r}\neq \emptyset$
	and $\tuple{\Ttrace,\Ttrace},i \models \bigwedge_{p\in \Head{r}\setminus L} \neg p$.
	So, $\tuple{\Ttrace,\Ttrace},i\not\models \bigvee_{a\in L} a \rightarrow \mathit{ES}_{\dynamic{P}}(L)$,
	and then $\tuple{\Ttrace,\Ttrace}\not\models \LF{P}$: a contradiction.\\

	\item For the case when $i=0$ we reach a contradiction in a similar way as above.
\end{enumerate}

\end{proofof}

\end{document}